\documentclass[11pt,a4paper]{article}
\usepackage[hyperref]{acl2021}
\usepackage{times}
\usepackage{latexsym}

\usepackage{microtype}

\aclfinalcopy 

\usepackage{graphicx}
\usepackage{amsmath}
\usepackage{subfigure}
\usepackage{multirow}

\title{RealFormer: Transformer Likes Residual Attention}

\author{Ruining He,  Anirudh Ravula, Bhargav Kanagal, Joshua Ainslie \\
Google Research \\
\texttt{\{ruininghe,braineater,bhargav,jainslie\}@google.com} }

\date{}

\begin{document}
\maketitle

\begin{abstract}
Transformer is the backbone of modern NLP models. In this paper, we propose \textbf{RealFormer}, a simple and generic technique to create \textbf{Re}sidual \textbf{A}ttention \textbf{L}ayer Trans\textbf{former} networks that significantly outperform the canonical Transformer and its variants (BERT, ETC, etc.) on a wide spectrum of tasks including Masked Language Modeling, GLUE, SQuAD, Neural Machine Translation, WikiHop, HotpotQA, Natural Questions, and OpenKP.
We also observe empirically that RealFormer stabilizes training and leads to models with \emph{sparser} attention.
Source code and pre-trained checkpoints for RealFormer can be
found at \url{https://github.com/google-research/google-research/tree/master/realformer}.
\end{abstract}

\section{Introduction}
\begin{figure*}[!h]
\centering
\includegraphics[width=\textwidth,height=\textheight,keepaspectratio]{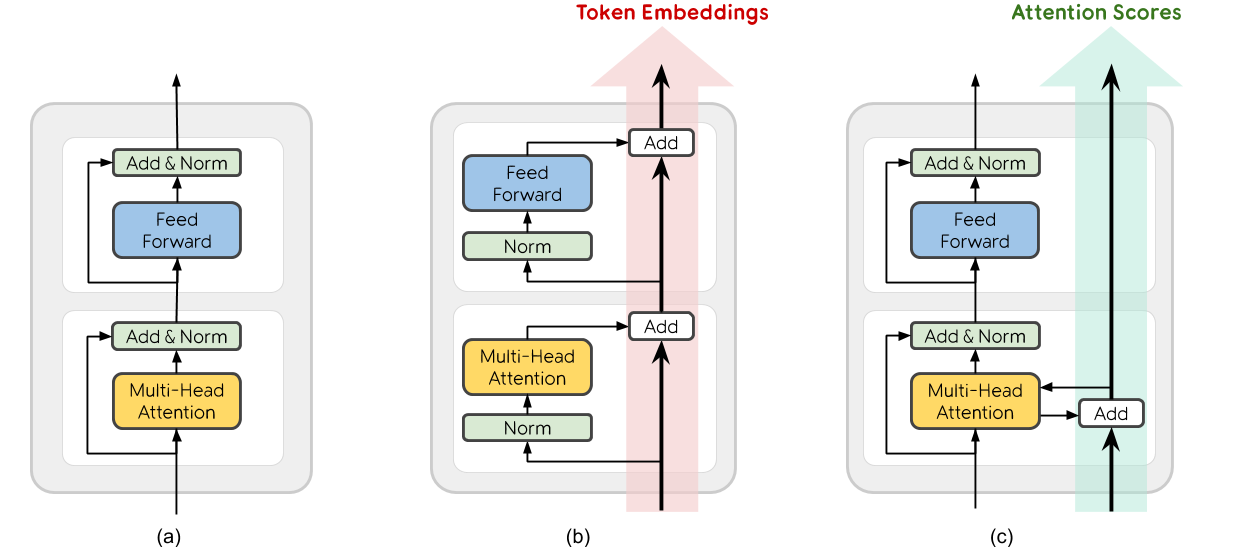}
\caption{Comparison of different styles of Transformer layers: (a) The prevalent Post-LN layer used by (\emph{e.g.}) BERT; (b) Pre-LN layer used by (\emph{e.g.}) GPT-2 that creates a ``direct'' path to propagate token embeddings; (c) Our RealFormer layer that creates a ``direct'' path to propagate attention scores (by adding a simple skip edge on top of (a)). Note that here we are showing Transformer encoder for demonstration purposes only; RealFormer can be applied straightforwardly for different Transformer variations (\emph{e.g.}, when decoders are involved).}
\label{fig:realformer}
\end{figure*}

Transformer~\citep{Vaswani-2017-attention} architectures are the backbone of numerous state-of-the-art NLP models such as BERT ~\citep{Devlin-2019-bert}, GPT~\citep{Radford-2019-gpt2}, and Meena~\citep{Adiwardana-2020-meena}, and have seen wide success across both academia and industry. Typically, a Transformer network consists of a stack of residual layers. The original design follows a ``Post-LN'' structure which adds Layer Norm (LN) as a ``post-processing'' step for each sub-layer, as shown in Figure~\ref{fig:realformer}~(a). It has been adopted by various state-of-the-art models including BERT, XLNet~\citep{Yang-2019-xlnet}, RoBERTa~\citep{Liu-2019-roberta}, ALBERT~\citep{Lan-2019-albert}, Transformer-XL~\citep{Dai-2019-transformerxl}, and ETC~\citep{Ainslie-2020-etc}. Another notable design is to reorganize the order of modules to create a ``direct''/clean path to propagate embeddings of tokens in the input sequence through the whole network, as shown in Figure~\ref{fig:realformer}~(b).\footnote{Note that a final LN module is usually added at the very top of the whole network.} This design adds LN as a ``pre-processing'' step for each sub-layer, and is often referred to as ``Pre-LN'' and used by some well-known extra large models such as GPT-2~\citep{Radford-2019-gpt2} and Megatron~\citep{Shoeybi-2019-megatron}. In some respect, Post-LN and Pre-LN are analogous to ResNet v1~\citep{He-2016-resnetv1} and ResNet~v2~\citep{He-2016-resnetv2} respectively in the Computer Vision literature. Although ResNet~v2 is usually preferable to v1 for Computer Vision, it does not appear to be the case for Pre-LN Transformer in the NLP literature. It is likely that the particularities of self-attention modules and Transformer architectures potentially favor (at least slightly) different designs compared to traditional convolutional neural networks. 

In this paper, we propose a simple and generic technique to show that it is beneficial to create a ``direct'' path to propagate raw attention scores through Transformer-based networks. Our technique is called \textit{\textbf{Re}sidual \textbf{A}ttention \textbf{L}ayer Trans\textbf{former}}, or \textit{\textbf{RealFormer}} in short. 
We also use \emph{RealFormer} to denote the resulting Transformer networks whenever no confusion may arise.
Without losing generality, taking the standard Transformer encoder as an example, each RealFormer layer takes the raw attention scores of all attention heads from the previous layer and adds ``residual scores'' (computed the same way as attention scores in regular Transformers) on top, as shown in Figure~\ref{fig:realformer}~(c). The sum of the two scores is then used to compute attention probabilities via softmax.

In other words, RealFormer can be seen as adding simple skip connections to a backbone Transformer. Since it does not add expensive multiplication ops, performance is expected to be comparable.\footnote{In certain settings, we find it helpful for RealFormer to use running \emph{mean} of attention scores instead of running \emph{sum}, though it adds some negligible amount of multiplications.}
Note that our technique can also be applied straightforwardly for different Transformer variations and even when decoders are involved.

Specifically, our main contributions include:
\begin{itemize}
\item We present RealFormer, a simple, generic, and cheap technique to improve Transformer-based networks. It adds no parameters or hyper-parameters, and usually takes no more than a few lines of code changes to implement.
\item We show that RealFormer can be used as a drop-in replacement of Transformer in BERT, outperforming both Post-LN and Pre-LN Transformers across a wide spectrum of model sizes for pre-training. In terms of fine-tuning, it even achieves competitive downstream results when pre-trained with only half the number of epochs of the baselines.
\item We further demonstrate the genericity of RealFormer by using it as a drop-in replacement of two recent state-of-the-art Transformer variation models: ADMIN~\citep{Liu-2020-admin} from the Neural Machine Translation (NMT) domain, and ETC~\citep{Ainslie-2020-etc} that extends Transformer to handle long and structured inputs. We show that RealFormer can improve these models significantly on various tasks and lead to new state-of-the-art results.
\item Qualitatively, we observe that attention in RealFormer tends to be \emph{sparser} and more correlated across layers compared to baselines, which we believe may have some regularization effects that could stabilize training and benefit fine-tuning.
\end{itemize}

\section{Related Work}
\citet{Vaswani-2017-attention} proposed Transformer initially for NMT and it has profoundly changed the NLP field ever since. 

\citet{Radford-2018-gpt1} demonstrated that generative pre-training of a Transformer-based language model (GPT) on a diverse corpus of unlabeled text can give large gains to downstream NLP tasks that suffer from scarce labeled data. Following this thread, \citet{Devlin-2019-bert} proposed to pre-train a \emph{bidirectional} Transformer encoder (BERT) with a novel Masked Language Modeling as the main optimization objective. Since then, advances on many NLP tasks have been dominated by the self-supervised general-purpose pre-training, task-specific fine-tuning paradigm.
Following BERT, there has been a large stream of work that explores better self-supervision objectives (\emph{e.g.},~\citet{Yang-2019-xlnet,Clark-2020-electra}), larger pre-training data and better hyper-parameters (\emph{e.g.},~\citet{Liu-2019-roberta}), model parameter sharing (\emph{e.g.},~\citet{Lan-2019-albert}), multi-task pre-training (\emph{e.g.},~\citet{Sun-2019-ernie2,Raffel-2019-t5}).
These efforts typically employ a Post-LN Transformer at their core. In this paper we adopt BERT to test different Transformer architectures because it is widely used and representative of this body of work.

Another notable thread of work focuses on improving the efficiency/scalability of Transformer. Typically, they try to reduce the \emph{quadratic} complexity of the self-attention mechanism with respect to sequence length via low-rank methods (\emph{e.g.},~\citet{Wang-2020-linformer}), fixed strided attention patterns (\emph{e.g.},~\citet{Child-2019-sparsetransformer}), learnable attention patterns (\emph{e.g.},~\citet{Kitaev-2020-reformer, Roy-2020-routingtransformer}), memory-based global \& local attention (\emph{e.g.},~\citet{Ainslie-2020-etc, Beltagy-2020-longformer, Zaheer-2020-bigbird}), and so on. 
These methods are particularly useful when dealing with long documents that go beyond the capacity of standard Transformer models. We would refer the reader to~\citet{Tay-2020-survey} for a detailed survey.
RealFormer is orthogonal to these methods as it focuses on improving various Transformer networks with an universal technique which can apply to these models as well. 
In this paper, we will use RealFormer to improve a state-of-the-art model, ETC~\citep{Ainslie-2020-etc}, from this line of work to demonstrate the universality of RealFormer.

Some recent work (\emph{e.g.},~\citet{Wang-2019-deeppreln, Xiong-2020-preln, Zhang-2018-fixup, Huang-2020-tfixup, Zhang-2019-mergedattn}) has studied normalization and parameter initialization schemes for Transformers, though most evaluations focus only on NMT to the best of our knowledge. 
In this strand, \citet{Liu-2020-admin} recently proposed ADMIN, which achieved state-of-the-art results on multiple popular NMT benchmarks. 
In this paper, we will take ADMIN as an example to (1) evaluate RealFormer in settings involving decoders, and (2) show that it is possible to apply RealFormer on top of this line of work.

\section{RealFormer}
\subsection{Standard Transformer}
There is an encoder and a decoder in Transformer~\citep{Vaswani-2017-attention}. Since they work in a similar way, here we only introduce the encoder and refer the reader to the original paper for complete details.

There are two sub-layers inside each layer of a Transformer encoder. The first sub-layer contains a Multi-Head Attention module that computes output embeddings of a set of queries ($Q$) by aggregating the embeddings ($V$) of a set of keys ($K$):
\begin{align*}
\text{MultiHead} \, & (Q,K,V) = \\ 
                    & \text{Concat} \, (head_1, ..., head_h)\, W^O,
\end{align*}
where $head_i = \text{Attention} \, (QW^Q_i, KW^K_i, VW^V_i)$. $Q$ and $K$ are matrices with dimension $d_k$ and $V$ is a matrix with dimension $d_v$. $W^Q_i$, $W^K_i$, and $W^V_i$ are matrices that linearly project queries, keys, and values into the ``attention space'' of the $i$-th head. $W^O$ is a matrix that linearly transforms the concatenation of the outputs of all heads.

The attention function is typically implemented with a Scaled Dot-Product Attention module~\citep{Vaswani-2017-attention} which computes a weighted sum of the values:
\begin{align*}
\text{Attention} \, (Q', K', V') = \text{Softmax} \, (\frac{Q'K'^T}{\sqrt{d_k}}) \, V',
\end{align*}
where matrix $\frac{Q'K'^T}{\sqrt{d_k}}$ contains the raw attention scores for each (query, key) pair. These scores are normalized via the Softmax function for each query and then act as weights for the corresponding vectors in $V'$.

The second sub-layer contains a fully-connected Feed-Forward Network (FFN) module with one hidden layer:
\begin{align*}
\text{FFN} \, (x) = \sigma \, (x \, W_1 + b_1) \, W_2 + b_2,
\end{align*}
where $\sigma$ is an activation function usually implemented with ReLU or GELU (\emph{e.g.},~\citet{Devlin-2019-bert}). FFN is applied to each position in the sequence separately and identically.
Finally, there are Layer Norm (LN) modules inserted into the above two sub-layers to stabilize training.

As shown in Figure~\ref{fig:realformer}, there are two canonical designs of the Transformer network which only differ in the ways they organize the modules. Post-LN is the original architecture proposed by~\citet{Vaswani-2017-attention} which normalizes the outputs at the end of each sub-layer. In contrast, Pre-LN normalizes sub-layer inputs instead and creates a direct path (without LN in the way) to propagate embeddings of the tokens in the sequence.

\subsection{Residual Attention Layer Transformer} \label{sec:realformer}
RealFormer uses a Post-LN style Transformer\footnote{Potentially we could also use Pre-LN, but Post-LN tends to outperform Pre-LN, as we will show in Section~\ref{sect:exp}.}  as backbone and adds skip edges to connect Multi-Head Attention modules in adjacent layers, as shown in Figure~\ref{fig:realformer} (c). More formally, it adds $Prev$, the pre-softmax attention scores from the previous layer with shape $(\#heads, from\_seq\_len, to\_seq\_len)$,\footnote{Batch dimension is omitted for ease of discussion.} as one additional input to the Multi-Head Attention module in the current layer:
\begin{align*}
\text{Residual} & \text{MultiHead} \, (Q, K, V, Prev) = \\ 
                & \text{Concat} \, (head_1, ..., head_h)\, W^O,
\end{align*}
where $head_i = \text{ResidualAttention} \, (QW^Q_i, $ $ KW^K_i, VW^V_i, Prev_i)$ and $Prev_i$ is the slice of $Prev$ with shape $( from\_seq\_len, to\_seq\_len)$ corresponding to $head_i$.
ResidualAttention adds ``residual scores'' on top of $Prev_i$ and then computes the weighted sum as usual:
\begin{equation} \label{eq:residual_attn}
\begin{split}
\text{Residual} & \text{Attention}  \, (Q', K', V', Prev') = \\  
                 & \text{Softmax} \, (\frac{Q'K'^T}{\sqrt{d_k}} + Prev') \, V'.
\end{split}
\end{equation}
Finally, new attention scores $\frac{Q'K'^T}{\sqrt{d_k}} + Prev'$ are passed over to the next layer.

Implementing RealFormer takes no more than adding a few lines of code to the backbone Transformer. Note that the RealFormer technique can be straightforwardly applied for Transformer variations and even when there are more than one type of attention modules in the network. For example, there are encoder self-attention, encoder-decoder attention, and decoder self-attention modules for machine translation. In such cases, RealFormer simply adds skip edges to create \emph{multiple} direct paths, one for each type of attention module.

\paragraph{Discussion.} Adding skip edges is equivalent to using a softmax over the running sum of the attention scores (to get attention probabilities). This might be sub-optimal for very deep networks due to the linear scaling nature of sum. Empirically, we find it helpful to use running \emph{mean} instead in such cases, which can be viewed as adding a temperature (\emph{i.e.}, \#traversed layers) to the softmax function in Eq.~\ref{eq:residual_attn} of each RealFormer layer.


\section{Experiments} \label{sect:exp}
To demonstrate that RealFormer is general-purpose, we conduct comprehensive empirical studies on a variety of tasks including (masked) language modeling, machine translation, and long document modeling, based on corresponding state-of-the-art models: BERT, ADMIN, and ETC.
To evaluate its robustness, we only do minimal (if at all) hyper-parameter tuning for RealFormer and initialize all parameters the same way as the backbone Transformers.
More aggressive hyper-parameter tuning or better initialization might further improve RealFormer, though we leave them for future work.
Details of our experiments are included in Appendix.

\subsection{BERT} \label{sec:bert}
BERT~\citep{Devlin-2019-bert} has been the standard way of transferring knowledge from large unlabeled text corpora by pre-training a bidirectional Transformer encoder. Numerous downstream NLP tasks suffering from scarcity of supervised data have benefited considerably by fine-tuning a pre-trained BERT model. This drives us to adopt BERT as the main evaluation setup for RealFormer.

\paragraph{Experiment setup.}
Our experiments are based on the official BERT repository\footnote{\url{https://github.com/google-research/bert}}. We follow the standard pre-training setup (dataset: Wikipedia + BookCorpus, vocab: uncased 30K, max sequence length: 512\footnote{Unlike BERT which uses a reduced sequence length for the first 90\% of steps, we always use 512 for simplicity.}, dropout: 10\%, learning rate: 1e-4, learning rate schedule: warm up and then linearly decay to 0, weight decay: 0.01, optimizer: AdamW, objective: Masked Language Modeling + Next Sentence Prediction, etc.) to compare three Transformer models: Post-LN, Pre-LN, and RealFormer. 
We experiment with Transformer architectures with a wide spectrum of sizes as detailed in Table~\ref{table:bert-arch}. For simplicity, all models are pre-trained 1M steps with a mini-batch size of 512 (except that xLarge uses 256 to avoid TPU OOM). Note that we use a larger mini-batch size than~\citet{Devlin-2019-bert}, \emph{i.e.}, doubling the amount of pre-training epochs, to show more complete behavior of different models.

\begin{table}
\setlength{\tabcolsep}{5pt}
\centering
\begin{tabular}{l|ccccc}
\hline \textbf{Model} & \textbf{L} & \textbf{H} & \textbf{A} & \textbf{I} & \textbf{P} \\ \hline
BERT-Small  & 4  & 512   & 8  & 2,048 & 30M  \\
BERT-Base   & 12 & 768   & 12 & 3,072 & 110M \\
BERT-Large  & 24 & 1,024 & 16 & 4,096 & 340M \\
BERT-xLarge & 36 & 1,536 & 24 & 6,144 & 1B \\
\hline
\end{tabular}
\caption{\label{table:bert-arch} Model architectures for BERT evaluation. L: \#layers, H: hidden size, A: \#heads, I: intermediate size, P: approximate \#parameters.}
\end{table}

\begin{figure*}
\subfigure[BERT-Small]{\includegraphics[width=1.61in, keepaspectratio]{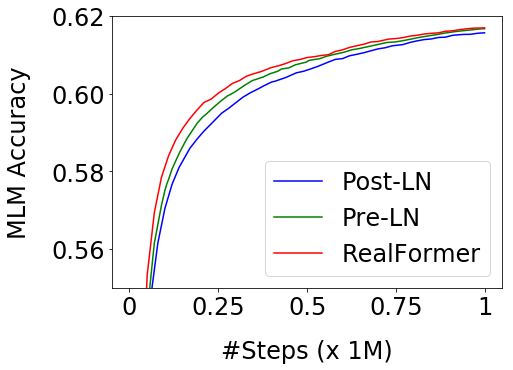}}
\subfigure[BERT-Base]{\includegraphics[trim=33 0 0 0,clip,width=1.52in,keepaspectratio]{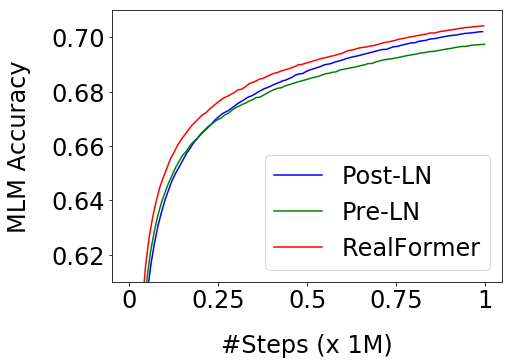}}
\subfigure[BERT-Large]{\includegraphics[trim=33 0 0 0,clip,width=1.52in,keepaspectratio]{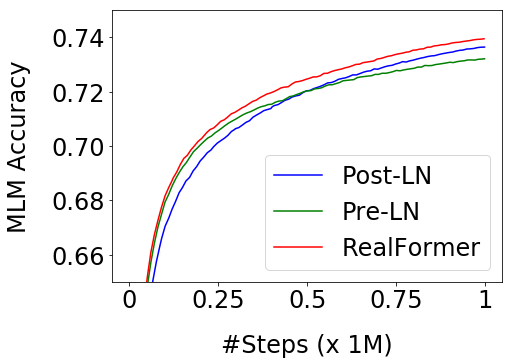}}
\subfigure[BERT-xLarge]{\includegraphics[trim=33 0 0 0,clip,width=1.52in,keepaspectratio]{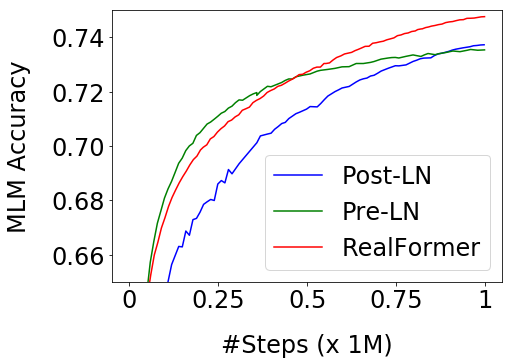}}
\caption{Development set MLM accuracy (best viewed in color). Improvement gap of RealFormer over the best baseline tends to increase with model size. Note that these are without hyper-parameter tuning for RealFormer. (As we will show later, RealFormer can benefit from larger learning rates and even double the gap size over Post-LN.)}
\label{fig:mlm}
\end{figure*}

We use exactly the same setup for all three Transformer architectures except that for the Pre-LN Transformer we follow the initialization strategy suggested by~\citet{Radford-2019-gpt2} and~\citet{Child-2019-sparsetransformer}.\footnote{We also tried BERT-style initialization in our pilot experiments but without success.} Note that for simplicity RealFormer reuses all hyper-parameter setups from Post-LN Transformer unless otherwise specified.
We use running sum of attention scores for all RealFormer models except xLarge (for which we use running mean for reasons discussed in Section~\ref{sec:realformer}).

All experiments are performed on 128 or 256 TPU v3 cores depending on model sizes (see Appendix~\ref{sec:appendix:bert} for details).

\begin{table}
\setlength{\tabcolsep}{3.5pt}
\centering
\begin{tabular}{l|ccc}
\hline \textbf{Model} & \textbf{Post-LN} & \textbf{Pre-LN} & \textbf{RealFormer} \\ \hline
BERT-Small  & 61.57\%  & 61.67\%   & \textbf{61.70\%}  \\
BERT-Base   & 70.20\%  & 69.74\%   & \textbf{70.42\%}  \\
BERT-Large  & 73.64\%  & 73.21\%   & \textbf{73.94\%}  \\
BERT-xLarge & 73.72\%  & 73.53\%   & \textbf{74.76\%}  \\
\hline
\end{tabular}
\caption{Development set MLM accuracy after pre-training 1M steps. RealFormer outperforms baselines more as model size increases.}
\label{table:bert-mlm}
\end{table}

\subsubsection{Pre-training Results} \label{sec:pre-train}
To evaluate pre-trained models, we report Masked Language Modeling (MLM) accuracy\footnote{All methods achieved similar (and great) results on Next Sentence Prediction presumably because it is much easier.} on a randomly held-out development set. As shown in Table~\ref{table:bert-mlm}, RealFormer outperforms the two baseline Transformers considerably with the gap increasing with model size. 
Our hypothesis is that larger models are inherently harder to train (\emph{e.g.}, we observe that BERT with Post-LN is unstable and sometimes even diverges for xLarge) and RealFormer can help regularize the model and stabilize training. 

We also report the pre-training curves in Figure~\ref{fig:mlm}. One interesting finding is that the Pre-LN Transformer seems to favor the combination of extra large models and a small number of steps, though it is consistently outperformed by the other two in ``regular-sized'' settings or given enough pre-training budget.

\subsubsection{Downstream Results}
To evaluate downstream performance, we fine-tune the above pre-trained BERT-Large models on both sentence-level (\emph{i.e.}, GLUE) and token-level (\emph{i.e.}, SQuAD) NLP tasks.

\paragraph{GLUE.}
General Language Understanding Evaluation (GLUE) is a canonical benchmark proposed by \citet{Wang-2018-glue} for evaluating models across a diverse set of NLU tasks. Following the fine-tuning recipe in~\citet{Devlin-2019-bert}, we use a mini-batch size of 32 for all models on all tasks. For each (task, model) pair, we select number of fine-tuning epochs in \{2, 3, 4\} and learning rate in \{6e-6, 8e-6, 1e-5, 2e-5, 3e-5, 4e-5, 5e-5\}.\footnote{We use a slightly wider range than~\citet{Devlin-2019-bert} to better accommodate all three models.} For each setup, we run the experiment five times and report the best median performance and the corresponding standard deviation on the development set.

\begin{table}[t]
\setlength{\tabcolsep}{3.5pt}
\centering
\begin{tabular}{l|ccc}
\hline 
\textbf{Task}      & \textbf{Post-LN}     & \textbf{Pre-LN}       & \textbf{RealFormer}      \\ \hline
MNLI-m             & 85.96\tiny{$\pm$0.11}  & 85.03\tiny{$\pm$0.12}   & \textbf{86.28}\tiny{$\pm$0.14}  \\
MNLI-nm            & 85.98\tiny{$\pm$0.14}  & 85.05\tiny{$\pm$0.19}   & \textbf{86.34}\tiny{$\pm$0.30}  \\
QQP                & 91.29\tiny{$\pm$0.10}  & 91.29\tiny{$\pm$0.16}   & \textbf{91.34}\tiny{$\pm$0.03}  \\
QQP \small{(F1)}   & \textbf{88.34}\tiny{$\pm$0.15}  & 88.33\tiny{$\pm$0.26}   & 88.28\tiny{$\pm$0.08}  \\
QNLI               & 92.26\tiny{$\pm$0.15}  & \textbf{92.35}\tiny{$\pm$0.26}   & 91.89\tiny{$\pm$0.17}  \\
SST-2              & 92.89\tiny{$\pm$0.17}  & 93.81\tiny{$\pm$0.13}   & \textbf{94.04}\tiny{$\pm$0.24}  \\
CoLA \small{(MC)}  & 58.85\tiny{$\pm$1.31}  & 58.04\tiny{$\pm$1.50}   & \textbf{59.83}\tiny{$\pm$1.06}  \\
STS-B \small{(PC)} & 90.08\tiny{$\pm$0.27}  & 90.06\tiny{$\pm$0.33}   & \textbf{90.11}\tiny{$\pm$0.56}  \\
STS-B \small{(SC)} & 89.77\tiny{$\pm$0.26}  & 89.62\tiny{$\pm$0.28}   & \textbf{89.88}\tiny{$\pm$0.54}  \\
MRPC               & \textbf{87.50}\tiny{$\pm$0.67}  & 86.76\tiny{$\pm$5.64}   & 87.01\tiny{$\pm$0.91}  \\
MRPC \small{(F1)}  & \textbf{91.16}\tiny{$\pm$0.45}  & 90.69\tiny{$\pm$3.16}   & 90.91\tiny{$\pm$0.65}  \\
RTE                & 71.12\tiny{$\pm$2.52}  & 68.59\tiny{$\pm$1.52}   & \textbf{73.65}\tiny{$\pm$0.90}  \\ \hline
Overall            & 84.01                  & 83.47                   & \textbf{84.53}  \\
\hline
\end{tabular}
\caption{\label{table:glue} GLUE development set results of fine-tuning BERT-Large models in Table~\ref{table:bert-mlm}. Default metric: accuracy, MC: Matthews correlation, PC: Pearson correlation, SC: Spearman correlation. Overall: first average metrics within each task (if there are 1+) and then across tasks. Numbers in smaller font are standard deviations. All numbers are scaled by 100.}
\end{table}
 
Results are tabulated in Table~\ref{table:glue}. We exclude the problematic WNLI task following~\citet{Devlin-2019-bert}. For each task, we report metric(s) suggested by \citet{Wang-2018-glue}. RealFormer achieves the best overall performance and outperforms both baselines on most tasks, testifying its strength at tackling sentence-level tasks.

\paragraph{SQuAD.}
The Stanford Question Answering Dataset (SQuAD v1.1) is a reading comprehension dataset consisting of 100K crowd-sourced question-answer pairs, where the answer to each question is a segment of text from the corresponding reading passage~\citep{Rajpurkar-2016-squad}.
SQuAD v2.0, a later version, further extends with over 50K unanswerable questions written adversarially by crowdworkers to look similar to answerable ones.

\begin{table}[t]
\setlength{\tabcolsep}{0.9pt}
\centering
\begin{tabular}{l|cccc}
\hline \textbf{SQuAD} &\textbf{Public}  & \textbf{Post-LN} & \textbf{Pre-LN} & \textbf{RealFormer} \\ \hline
v1.1 \small{(F1)}   & 90.9  & 91.68\tiny{$\pm$0.12}  & 91.06\tiny{$\pm$0.09}    & \textbf{91.93}\tiny{$\pm$0.12}   \\
v1.1 \small{(EM)}   & 84.1  & 85.15\tiny{$\pm$0.13}  & 83.98\tiny{$\pm$0.24}    & \textbf{85.58}\tiny{$\pm$0.15}   \\
v2.0 \small{(F1)}   & 81.9  & 82.51\tiny{$\pm$0.12}  & 80.30\tiny{$\pm$0.12}    & \textbf{82.93}\tiny{$\pm$0.05}   \\
v2.0 \small{(EM)}   & 78.7  & 79.57\tiny{$\pm$0.12}  & 77.35\tiny{$\pm$0.16}    & \textbf{79.95}\tiny{$\pm$0.08}  \\
\hline
\end{tabular}
\caption{\label{table:squad} SQuAD development set results of fine-tuning BERT-Large models in Table~\ref{table:bert-mlm}. EM: exact match. Public: Post-LN results from~\citet{Devlin-2019-bert}. Numbers in smaller font are standard deviations. All numbers are scaled by 100.}
\end{table}

We follow the fine-tuning recipe in~\citet{Devlin-2019-bert} for all three Transformer models on these two datasets without using any additional data such as TriviaQA~\citep{Joshi-2017-triviaqa}.
For both v1.1 and v2.0, we select mini-batch size in \{32, 48\}, number of fine-tuning epochs in \{2, 3, 4\}, and learning rate in \{2e-5, 3e-5, 4e-5, 5e-5\}. For each setup, we run the experiment five times and report the best median performance and the corresponding standard deviation on the development set. As we can see from Table~\ref{table:squad}, RealFormer outperforms the two baselines considerably, attesting its strength at tackling token-level tasks.

\subsubsection{Research Questions} \label{sec:rq}

\paragraph{How well does RealFormer perform with half the pre-training budget?}
Although RealFormer has outperformed both Post-LN and Pre-LN considerably when pre-training 1M steps, we are also interested in investigating its potential when the pre-training budget is more limited. For this purpose, we experiment with BERT-Large models. In particular, we take the 500K step checkpoint of the pre-trained RealFormer in Table~\ref{table:bert-mlm} and fine-tune it on GLUE and SQuAD datasets using exactly the same procedure as described above. Comparison results against the strongest baseline, Post-LN Transformer pre-trained 500K (checkpoint) and 1M steps respectively, are collected in Table~\ref{table:500k}. We can see that RealFormer with merely half the amount of pre-training epochs can beat Post-LN (1M) on GLUE with a significant margin, and almost match its performance on SQuAD.

\begin{table}
\setlength{\tabcolsep}{2.8pt}
\centering
\begin{tabular}{l|ccc}
\hline 
\multirow{2}{*}{\textbf{Task}}   & \multicolumn{1}{p{1.75cm}}{\centering \textbf{Post-LN} \\ \small{(500K)}}      & \multicolumn{1}{p{1.75cm}}{\centering \textbf{Post-LN} \\ \small{(1M)}}      & \multicolumn{1}{p{2cm}}{\centering \textbf{RealFormer} \\ \small{(500K)}} \\ \hline
GLUE                   & 83.84                   & 84.01                   & 84.34       \\
v1.1 \small{(F1)}      & 91.46\tiny{$\pm$0.18}   & 91.68\tiny{$\pm$0.12}   & 91.56\tiny{$\pm$0.09}       \\
v1.1 \small{(EM)}      & 84.87\tiny{$\pm$0.24}   & 85.15\tiny{$\pm$0.13}   & 85.06\tiny{$\pm$0.12}       \\
v2.0 \small{(F1)}      & 81.44\tiny{$\pm$0.50}   & 82.51\tiny{$\pm$0.12}   & 82.52\tiny{$\pm$0.55}       \\
v2.0 \small{(EM)}      & 78.64\tiny{$\pm$0.48}   & 79.57\tiny{$\pm$0.12}   & 79.54\tiny{$\pm$0.54}       \\ \hline
Overall                & 83.97                   & 84.37                   & \textbf{84.51}    \\ \hline
\end{tabular}
\caption{\label{table:500k} Downstream development set results of fine-tuning BERT-Large with Post-LN and RealFormer pre-trained with different number of steps. v*: SQuAD version, EM: exact match. Overall: First average across SQuAD and then GLUE. Numbers in smaller font are standard deviations. All numbers are scaled by 100.}
\end{table}

\begin{figure}
\centering
\includegraphics[width=0.475\textwidth,keepaspectratio]{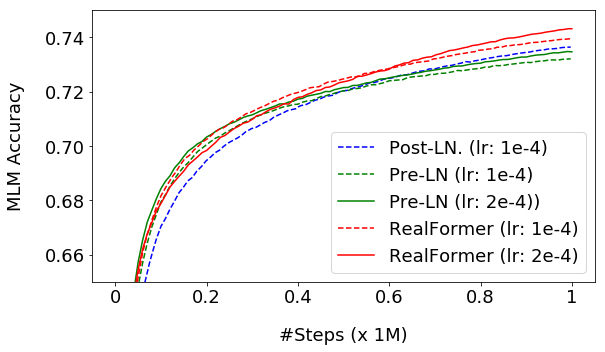}
\caption{Development set MLM accuracy of BERT-Large with different learning rates (best viewed in color). RealFormer seems to benefit slightly more from using a larger, non-default learning rate compared to Pre-LN, while Post-LN diverges with 2e-4.}
\label{fig:lr}
\end{figure}

\paragraph{Does a larger learning rate help?}
As suggested by some recent work (\emph{e.g.},~\citet{Xiong-2020-preln}), Pre-LN Transformer may benefit from using larger learning rates. To this end, we follow the pre-training procedure detailed earlier and switch to a larger learning rate, 2e-4, to pre-train BERT-Large with the three Transformer models. Development set MLM accuracy with training steps can be found in Figure~\ref{fig:lr}. We find that both Pre-LN and RealFormer can reap some benefits of using larger learning rates with RealFormer seeming to benefit slightly more in this case (73.94\% $\rightarrow$ 74.31\%) compared to Pre-LN (73.21\% $\rightarrow$ 73.46\%). Post-LN diverges with the learning rate of 2e-4.
Note that it also means that RealFormer can outperform Post-LN, the strongest baseline, actually with a prominent gap, 0.67\% (\emph{i.e.}, 74.31\% - 73.64\%), for pre-training, though with only minimal learning rate tuning.

\paragraph{Is attention sparser in RealFormer?}
We conduct one empirical study to observe the qualitative differences between RealFormer and Post-/Pre-LN Transformers. 
We randomly sample 8,192 examples from the held-out development set and visualize the distribution of attention probabilities of each token in these examples across heads in all layers.
In particular, for each (token, layer, head) triplet, we compute the entropy of the attention probabilities as the ``sparsity measure'' of attention. Intuitively, as entropy gets lower, the attention weight distribution becomes more skewed and therefore attention is sparser.

\begin{figure*}[!t]
\centering
\includegraphics[width=\textwidth,keepaspectratio]{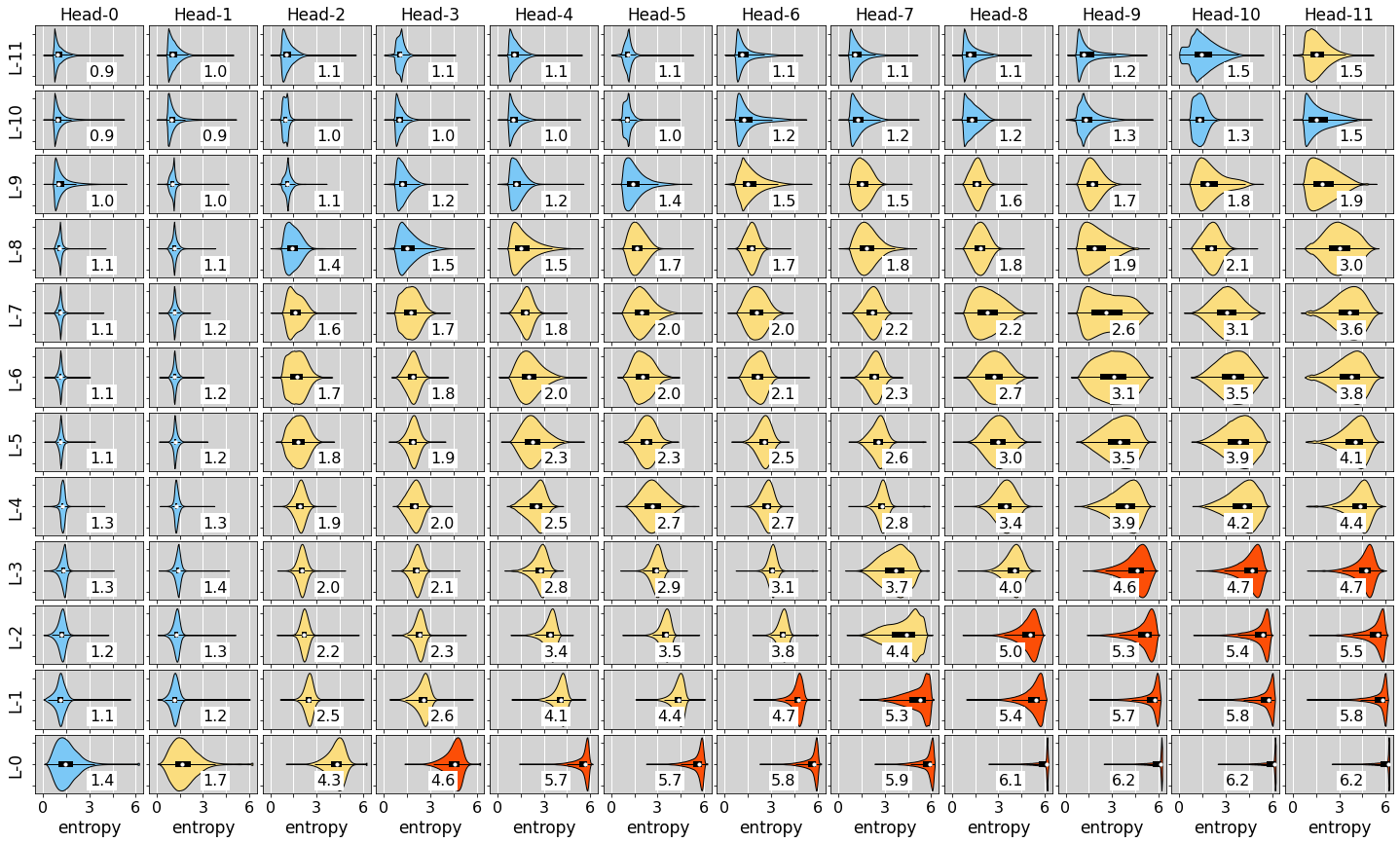}
\caption{Distribution of entropies of the attention probabilities of the tokens of 8,192 held-out examples using the pre-trained BERT-Base with \textbf{RealFormer} (see Section~\ref{sec:pre-train}). For better legibility, (1) attention heads in each layer are ordered by their medians of entropies, and (2) distributions are color-coded based on the median of entropies: RED (median $>$ 4.5), YELLOW (1.5 $\le$ median $\le$ 4.5), BLUE (median $<$ 1.5), \emph{i.e.}, colder colors mean sparser attention. There is a clear trend that higher layers tend to have \emph{sparser} attention.}
\label{fig:realformer-entropy}
\end{figure*}

In a similar fashion to~\citet{Ramsauer-2020-hopfield}, we use violin plots to show the entropy distributions of the pre-trained BERT-Base model with RealFormer from Table~\ref{table:bert-mlm} (see Figure~\ref{fig:realformer-entropy}).
Plots for the two baseline Transformers in Table~\ref{table:bert-mlm} are included in Appendix~\ref{sec:appendix:entropy}.
Each row is a layer in BERT-Base and each column is an attention head.

 We find that attention tends to get sparser for later (upper) layers for all three Transformers. However, RealFormer differs from the two baselines in the following ways: 
\begin{itemize}
    \item RealFormer has \emph{significantly sparser} attention for top layers (layer 9-11);
    \item RealFormer tends to have lower variance across all layers, which means that attention density is less input-dependent.
\end{itemize}
We hypothesize that the above two properties might be a sign of stableness and benefit fine-tuning.

\paragraph{Do attention heads in layer $L$ resemble those in layer $L-1$?}
Since RealFormer uses a residual attention scheme, it is interesting to show to what extent an attention head is ``relying on'' the corresponding head in the previous layer.
To this end, we take each of the three pre-trained BERT-Base models in Table~\ref{table:bert-mlm} and compute the Jensen-Shannon Divergence (JSD) between attention probabilities in each pair of \emph{vertically} adjacent heads, \emph{i.e.}, $\text{JSD}\,(\text{head}^L_i, \text{head}^{L-1}_i)$, for $1 \le L < 12$ and $0 \le i < 12$.

Appendix~\ref{sec:appendix:jsd} demonstrates detailed JSD distributions of Post-LN
and RealFormer respectively based on 8,192 held-out examples. We observe that RealFormer tends to have \emph{significantly lower} JSD values (\emph{i.e.}, indicating more ``similar'' attention across layers), especially for heads in middle layers. This might mean that RealFormer has some regularization advantages and provides one hypothesis for why it tends to outperform Post-LN more for larger models. Note that $\text{head}^L_i$ can still be useful even if it has exactly the same attention probabilities with $\text{head}^{L-1}_i$ because of the existence of the FFN sub-layer and the potential differences in value matrices (\emph{i.e.}, $V'$ in Eq.~\ref{eq:residual_attn}).

\paragraph{Is residual attention really necessary?}
One may wonder whether increasing dropout rate can already regularize large models well so that residual attention is redundant. To this end, we experiment with different dropout rates for pre-training BERT-Large with different Transformers (following the procedures in Section~\ref{sec:pre-train}). Results are collected in Table~\ref{table:dropout}, from which we can see that (1) RealFormer outperforms the two baselines across all dropout settings, and (2) simply increasing dropout rate can not regularize Transformer models as well as what residual attention appears to be doing.

\begin{table}
\centering
\begin{tabular}{l|ccc}
\hline \textbf{Dropout} & \textbf{Post-LN} & \textbf{Pre-LN} & \textbf{RealFormer} \\ \hline
0\%\footnotemark   & 71.16\%  & 69.80\%  & \textbf{71.30\%}  \\
10\%  & 73.64\%  & 73.21\%   & \textbf{73.94\%}  \\
20\%  & 73.21\%  & 72.97\%   & \textbf{73.66\%}  \\
\hline
\end{tabular}
\caption{Development set MLM accuracy of BERT-Large with different dropout rates.}
\label{table:dropout}
\end{table}

\footnotetext{When dropout rate is 0\%, we use early stopping for all models due to overfitting.}

\subsection{ADMIN}
To evaluate the genericity of RealFormer, here we try it on top of ADMIN~\citep{Liu-2020-admin}, a state-of-the-art NMT model without using either additional data or data augmentation. 
ADMIN adopts Post-LN as the backbone, which we simply replace with RealFormer.
In particular, we add three types of skip edges for encoder-encoder, encoder-decoder, and decoder-decoder attention respectively to the Post-LN Transformer. 
Empirically, RealFormer with running mean of attention scores tends to outperform running sum for our experiments, therefore here we use the former exclusively for brevity.

We use two popular NMT benchmarks, WMT'14 En-De and WMT'14 En-Fr, and follow \citet{Liu-2020-admin} for all training setups on both benchmarks except that in all cases (1) we select the peak learning rate from \{5e-4, 1e-3, 1.2e-3\} and use a linear learning rate decay schedule (instead of inverse sqrt);\footnote{With inverse sqrt decay, we find that RealFormer tends to favor larger peak learning rates than what~\citet{Liu-2020-admin} uses, and we have also seen improvements in most cases.} (2) we train RealFormer only 50 epochs (in contrast, ADMIN trains 100 epochs on En-De and 50 epochs on En-Fr); and (3) we average across the last 25 checkpoints (while ADMIN uses the last 10). 
More checkpoints are helpful for us (especially for large models) presumably because the last few are not ``diverse'' enough as learning rate decays to 0.

Our experiments are performed on NVIDIA A100 GPUs, based on the official ADMIN repository\footnote{\url{https://github.com/LiyuanLucasLiu/Transformer-Clinic}}.
We follow~\citet{Liu-2020-admin} to configure the amount of GPUs to use for different setups.

BLEU scores on test sets are collected in Table~\ref{table:nmt}. For fair comparisons, we also run ADMIN using our above setups and report results in the same table. Following~\citet{Liu-2020-admin}, all networks (including both encoders and decoders) share the same width setup (hidden size 512, intermediate size 2048, 8 heads) and only vary in depth. RealFormer outperforms all baselines across all depths considerably with a new state-of-the-art BLEU score (43.97) on En-Fr for models not using additional data or data augmentation to the best of our knowledge.
One interesting observation here is that RealFormer does not always lead to larger improvement gaps for larger models, which might be due to the checkpoint averaging mechanism (which potentially regularizes large models reasonably well).

\begin{table}
\setlength{\tabcolsep}{2.7pt}
\centering
\begin{tabular}{l|ccc|cccc} \hline
\multirow{2}{*}{\textbf{Model}}    &\multicolumn{3}{c|}{\textbf{En-De}}     &\multicolumn{2}{c}{\textbf{En-Fr}} \\
                          &\small{6L-6L} &\small{12L-12L} &\small{18L-18L}  &\small{6L-6L} &\small{60L-12L}     \\ \hline
Post-LN                   &27.80    &failed     &failed          &41.29   &failed    \\
Pre-LN                    &27.27    &28.26      &28.38           &40.74   &43.10     \\
ADMIN                     &27.90    &28.58      &29.03           &41.47   &43.80     \\ \hline
ADMIN$^\dagger$           &28.06    &28.85      &29.11           &41.65   &43.72     \\
Ours                      &\textbf{28.17} &\textbf{29.06}        &\textbf{29.35} &\textbf{41.92}   &\textbf{43.97}     \\  \hline
\end{tabular}
\caption{Test set BLEU scores on two WMT'14 benchmarks using different sizes of models. xL-yL: \#Encoder layers-\#Decoder layers. First three rows are from~\citet{Liu-2020-admin}. Ours is switching the backbone of ADMIN from Post-LN to RealFormer. $^\dagger$Our run of ADMIN using the same setups as RealFormer.}
\label{table:nmt}
\end{table}

\subsection{ETC}
Extended Transformer Construction (ETC) is a recent sparse attention mechanism proposed by~\citet{Ainslie-2020-etc} and \citet{Zaheer-2020-bigbird} to handle long context. It has achieved state-of-the-art results on four natural language benchmarks requiring long and/or structured inputs. Here we evaluate RealFormer on top of ETC models on these benchmarks including WikiHop~\citep{Welbl-2018-wikihop}, HotpotQA~\citep{Yang-2018-hotpotqa}, Natural Questions~\citep{Kwiatkowski-2019-nq}, and OpenKP~\citep{Xiong-2019-openkp}. 
They vary significantly in terms of dataset size, context length, and structure in
text inputs. Please refer to~\citet{Ainslie-2020-etc} for more details.

Our experiments are based on the official ETC repository\footnote{\url{https://github.com/google-research/google-research/tree/master/etcmodel}}. We take the ETC-Large model (24 layers, 1024 hidden size, 16 heads), add residual attention edges (\emph{i.e.}, using running sum), and follow all the pre-training and fine-tuning recipes as well as hardware setups detailed in~\citet{Ainslie-2020-etc}.
For each fine-tuning setup, we run the experiment five times and report the best median performance and the corresponding standard deviation on the development set in Table~\ref{table:etc}.
RealFormer can improve ETC consistently across all four benchmarks.

As of June 2021, we are ranked the first on the WikiHop leaderboard\footnote{\url{http://qangaroo.cs.ucl.ac.uk/leaderboard.html}}
with a test accuracy of 84.4\% (2.1\% absolute improvement over the previous best result).

\begin{table}
\setlength{\tabcolsep}{3.5pt}
\centering
\begin{tabular}{l|l||ccc} \hline
\textbf{Task}              & \textbf{Metric}    & \textbf{ETC}           &\textbf{Ours} \\ \hline
WikiHop                    & Accuracy           &78.92\tiny{$\pm$0.14}   &\textbf{79.21}\tiny{$\pm$0.38}   \\ \hline

\multirow{3}{*}{HotpotQA}  & Ans. F1            &80.38\tiny{$\pm$0.13}  &\textbf{80.86}\tiny{$\pm$0.16}   \\
                           & Sup. F1            &89.07\tiny{$\pm$0.06}  &\textbf{89.21}\tiny{$\pm$0.12}   \\
                           & Joint F1           &73.12\tiny{$\pm$0.19}  &\textbf{73.57}\tiny{$\pm$0.19}   \\ \hline

\multirow{3}{1.5cm}{Natural Questions}        & Long Ans. F1       &77.70\tiny{$\pm$0.15}   &\textbf{77.93}\tiny{$\pm$0.31}   \\
                           & Short Ans. F1      &58.54\tiny{$\pm$0.41}   &\textbf{59.10}\tiny{$\pm$0.81}   \\
                           & Average F1         &68.07\tiny{$\pm$0.17}   &\textbf{68.51}\tiny{$\pm$0.56}   \\ \hline

OpenKP                     & F1@3               &44.06\tiny{$\pm$0.08}   &\textbf{44.27}\tiny{$\pm$0.08}   \\ \hline
\end{tabular}
\caption{Development set results of ETC-Large. Ours is adding residual attention edges to ETC. Numbers in smaller font are standard deviations. All numbers are scaled by 100.}
\label{table:etc}
\end{table}


\section{Conclusions}
We propose RealFormer, a simple, generic, and cheap technique based on the novel idea of residual attention to improve Transformer-based networks.
Quantitatively, we show that RealFormer can improve a diverse set of state-of-the-art Transformer-based models considerably for tasks like Masked Language Modeling, Neural Machine Translation, and long document modeling.
Qualitatively, we show that RealFormer tends to have comparatively \emph{sparser} attention, both within heads and across heads in adjacent layers.

\bibliography{acl2021}
\bibliographystyle{acl_natbib}

\newpage

\appendix

\section{Appendices} 
\label{sec:appendix}

\subsection{Training Details: BERT}  \label{sec:appendix:bert}
All our experiments are conducted on TPUs based on \url{https://github.com/google-research/bert}, the official BERT repository in TensorFlow~\cite{Abadi-2016-tensorflow}.

\paragraph{Pre-training.} 
We use 128 TPU v3 cores (\emph{i.e.}, 64 chips) for BERT-Small/Base/Large and 256 TPU v3 cores (\emph{i.e.}, 128 chips) for BERT-xLarge. 
Table~\ref{table:bert-time} demonstrates the time used to pre-train each model 1M steps. We can see that there is 10\%-15\% performance drop when adding residual attention edges for all sizes except xLarge.
Our suspicion is that additions are not as optimized as other ops like matrix multiplications on TPU v3 cores.
There is a much smaller performance drop for xLarge though, which might indicate that addition scales nicely compared to other ops on TPU v3 cores.
As we will show later in Appendix~\ref{sec:appendix:admin}, performance drop on GPUs is almost negligible across different Transformer sizes, suggesting that it is hardware-dependent.

\begin{table}
\setlength{\tabcolsep}{3.5pt}
\centering
\begin{tabular}{l|ccc}
\hline \textbf{Model} & \textbf{Post-LN} & \textbf{Pre-LN} & \textbf{RealFormer} \\ \hline
BERT-Small  &5.4 hrs   &5.3 hrs    &5.9 hrs  \\
BERT-Base   &20 hrs    &20 hrs     &23 hrs   \\
BERT-Large  &58 hrs    &58 hrs     &66 hrs   \\
BERT-xLarge &136 hrs   &137 hrs    &137 hrs  \\
\hline
\end{tabular}
\caption{Pre-training time of different BERT models in Table~\ref{table:bert-mlm}. We use 128 TPU v3 cores and mini-batch size 512 for BERT-Small/Base/Large, and 256 TPU v3 cores and mini-batch size 256 for BERT-xLarge.}
\label{table:bert-time}
\end{table}

\begin{table}[t]
\setlength{\tabcolsep}{5pt}
\centering
\begin{tabular}{l|ccc|ccc}
\hline 
\multirow{2}{*}{\textbf{Task}}  &\multicolumn{3}{c|}{\textbf{500K-step}}     &\multicolumn{3}{c}{\textbf{1M-step}} \\
                                &BS   &LR      &EP                           &BS   & LR       &EP   \\ \hline
MNLI                            &32   &2e-5    &2                            &32   &1e-5      &4   \\  
QQP                             &32   &3e-5    &4                            &32   &2e-5      &4   \\  
QNLI                            &32   &3e-5    &4                            &32   &2e-5      &2   \\  
SST-2                           &32   &3e-5    &2                            &32   &1e-5      &4   \\  
CoLA                            &32   &2e-5    &4                            &32   &1e-5      &3   \\  
STS-B                           &32   &2e-5    &3                            &32   &2e-5      &4   \\  
MRPC                            &32   &2e-5    &4                            &32   &1e-5      &4   \\  
RTE                             &32   &1e-5    &4                            &32   &1e-5      &4   \\  \hline  
SQuAD v1.1                      &48   &3e-5    &2                            &48   &3e-5      &2   \\
SQuAD v2.0                      &32   &5e-5    &2                            &48   &5e-5      &2   \\ \hline
\end{tabular}
\caption{Hyper-parameter configurations on GLUE and SQuAD for best-performing BERT-Large with RealFormer (pre-trained 500K steps and 1M steps respectively). BS: mini-batch size, LR: learning rate, EP: \#fine-tuning epochs.}
\label{table:bert-hyperpara}
\end{table}

\begin{table}[h]
\setlength{\tabcolsep}{2.5pt}
\centering
\begin{tabular}{l|ccc|cccc} \hline
\multirow{2}{*}{\textbf{Model}}    &\multicolumn{3}{c|}{\textbf{En-De}}     &\multicolumn{2}{c}{\textbf{En-Fr}} \\
                          &\small{6L-6L} &\small{12L-12L} &\small{18L-18L}  &\small{6L-6L} &\small{60L-12L}     \\ \hline
ADMIN                     &9.3 hrs  &16 hrs   &23 hrs  &28 hrs   &70 hrs     \\
Ours                      &9.4 hrs  &16 hrs   &23 hrs  &28 hrs   &72 hrs    \\  \hline
\#GPUs                    &4 &4  &4 &8   &16     \\ \hline
\end{tabular}
\caption{Number of GPUs and training time used for each model. Ours is switching the backbone of ADMIN from Post-LN to RealFormer (using running mean of attention scores).}
\label{table:admin-time}
\end{table}

\paragraph{Fine-tuning.} 
We use 8 TPU v2 cores (\emph{i.e.}, 4 chips) to fine-tune each model.
Best hyper-parameter configurations for BERT-Large with RealFormer on GLUE and SQuAD are collected in Table~\ref{table:bert-hyperpara}. 
We include RealFormer pre-trained both 1M and 500K steps, corresponding to the results in Table~\ref{table:glue}, \ref{table:squad}, and \ref{table:500k}.

\begin{table*}[!h]
\setlength{\tabcolsep}{5pt}
\centering
\begin{tabular}{l|cccc|ccc}
\hline 
\multirow{2}{*}{\textbf{Task}}   &\multicolumn{4}{c|}{\textbf{Instance Statistics}}    &\multicolumn{3}{c}{\textbf{Hyper-parameter}}  \\
                                &\#Training  &Median Length  &95\%tile Length  &Max Length    &{BS}  &{LR}   &{EP}  \\ \hline
WikiHop                         &43,738      &1,541          &3,994            &20,337        &32    &4e-5   &15    \\ 
HotpotQA                        &90,447      &1,227          &1,810            &3,560         &32    &4e-5   &5     \\ 
Natural Questions               &307,373     &4,004          &17,137           &156,551       &64    &3e-5   &2     \\ 
OpenKP                          &133,724     &761            &4,546            &89,183        &64    &3e-5   &2     \\ \hline 
\end{tabular}
\caption{Statistics of benchmarks and the hyper-parameter configurations for best-performing ETC-Large with RealFormer. BS: mini-batch size, LR: learning rate, EP: \#fine-tuning epochs.}
\label{table:etc-hyperpara}
\end{table*}

\subsection{Training Details: ADMIN}  \label{sec:appendix:admin}
All our NMT experiments are conducted on NVIDIA A100 GPUs based on \url{https://github.com/LiyuanLucasLiu/Transformer-Clinic}, the official ADMIN repository implemented via fairseq~\citep{Ott-2019-fairseq}.
We use the same scripts to collect and process data and evaluate all models.

Following \citet{Liu-2020-admin}, we use different number of GPUs for different setups, as detailed in Table~\ref{table:admin-time}.
For our runs of ADMIN and RealFormer (\emph{i.e.}, the last two rows in Table~\ref{table:nmt}), learning rate is set to 5e-4 on WMT'14 En-De and 1.2e-3 on WMT'14 En-Fr across different model sizes.
We train all models 50 epochs across the two benchmarks and average across the last 25 checkpoints (corresponding to the last 25 epochs).

Training time comparison of ADMIN and our model using the same setups is shown in Table~\ref{table:admin-time}. 
Adding residual attention edges and using running mean of attention scores do not incur significant performance drop on GPUs across different model sizes.

\subsection{Training Details: ETC}
All our experiments are conducted on TPU v3 cores based on the official ETC repository in TensorFlow: \url{https://github.com/google-research/google-research/tree/master/etcmodel}.

\paragraph{Pre-training.} 
As is the case with ETC-Large \citep{Ainslie-2020-etc}, we find that pre-training ETC-Large with RealFormer can also benefit significantly from lifting weights from RoBERTa~\citep{Liu-2019-roberta}.
Note however that we lift from the same RoBERTa checkpoint as our ETC-Large baseline, which could be disadvantageous to our model since RoBERTa is pre-trained \emph{without} residual attention.

\paragraph{Fine-tuning.} 
Statistics of the four benchmarks and the corresponding best hyper-parameter configurations for ETC-Large with RealFormer are collected in Table~\ref{table:etc-hyperpara}. 
On Natural Questions and OpenKP, we simply reuse the best configurations for our ETC-Large baselines as reported in~\citet{Ainslie-2020-etc}. 
On WikiHop and HotpotQA, we follow the hyper-parameters search space specified in~\citet{Ainslie-2020-etc} for ETC-Large.\footnote{On WikiHop, number of fine-tuning epochs is selected from \{5, 10, 15\} instead of \{5, 10\} for both ETC-Large and our model. We added 15 here following the official ETC repository.} In addition, on WikiHop we found it to be slightly better (development set accuracy 79.21 vs 78.96) to turn off RealFormer during fine-tuning (\emph{i.e.}, adding no residual attention but still loading from our pre-trained RealFormer checkpoint); therefore we adopted this setup for WikiHop in Table~\ref{table:etc} and our leaderboard submission.

\subsection{Entropy Distribution of Pre-trained Baseline Transformer Models} \label{sec:appendix:entropy}

\begin{figure*}[!t]
\centering
\subfigure[Post-LN]{\includegraphics[width=\textwidth, keepaspectratio]{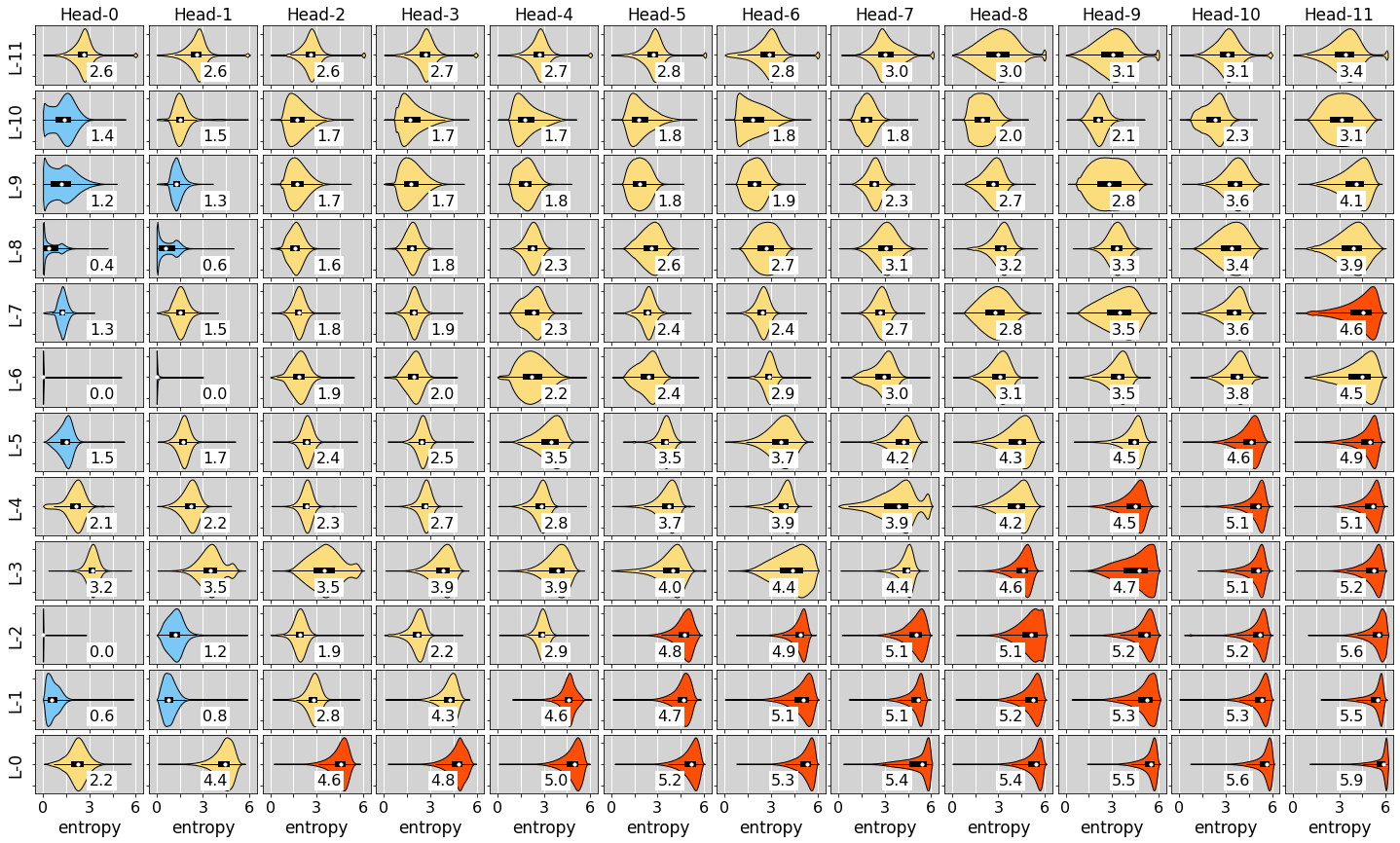}}
\subfigure[Pre-LN]{\includegraphics[width=\textwidth,keepaspectratio]{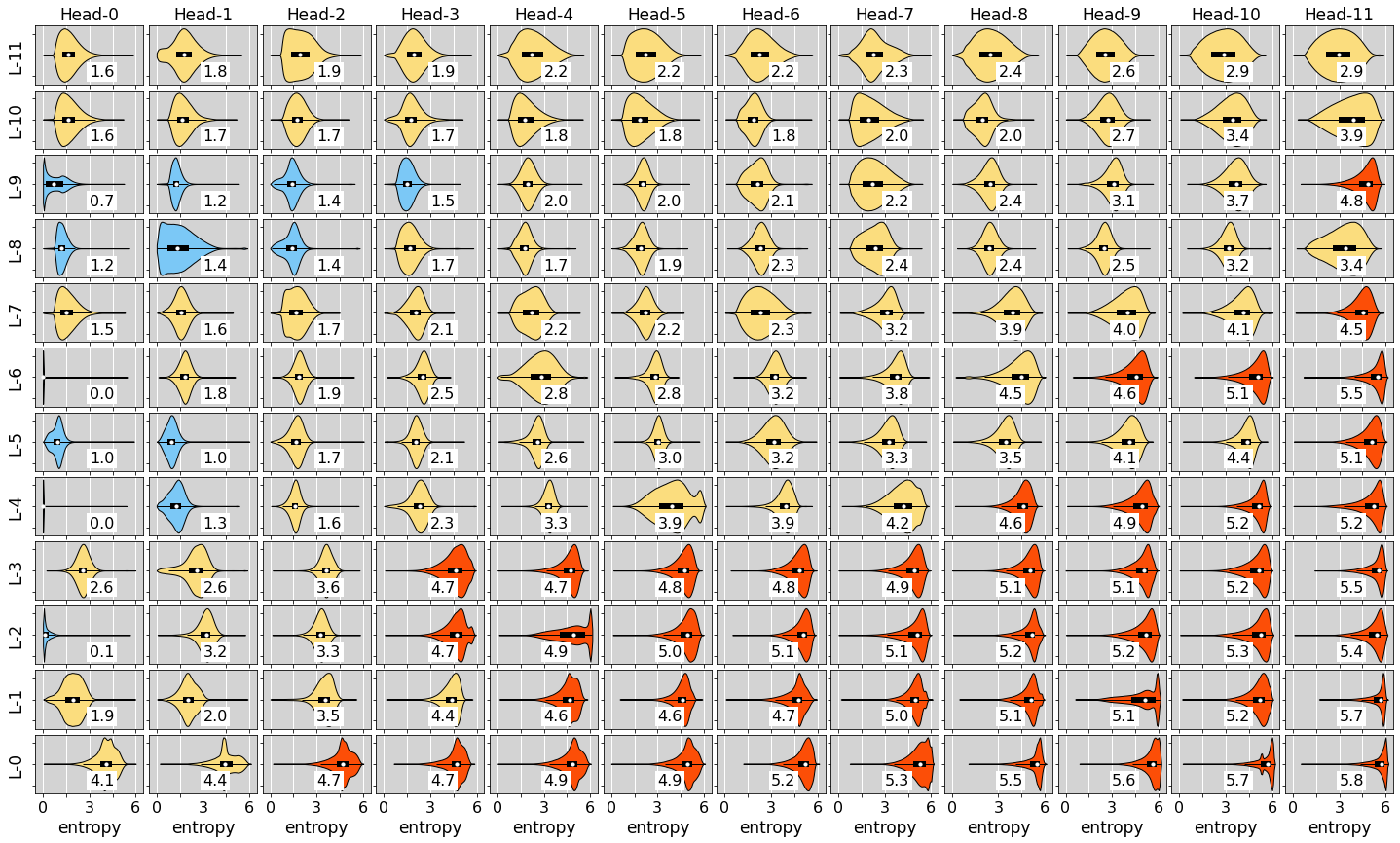}}
\caption{Distribution of entropies of the attention probabilities of the tokens of 8,192 held-out examples using the pre-trained BERT-Base with \textbf{Post-LN} and \textbf{Pre-LN} Transformer respectively (see Section~\ref{sec:pre-train}). For better legibility, (1) attention heads in each layer are ordered by their medians of entropies, and (2) distributions are color-coded based on the median of entropies: RED (median $>$ 4.5), YELLOW (1.5 $\le$ median $\le$ 4.5), BLUE (median $<$ 1.5), \emph{i.e.}, colder colors mean sparser attention. Note that here top layers (layer 9-11) tend to have larger entropies compared to RealFormer, which means that attention is relatively \emph{denser}.}
\label{fig:baseline-entropy}
\end{figure*}

Violin plots demonstrating the entropy distributions of the pre-trained BERT-Base models with Post-LN and Pre-LN Transformers from Table~\ref{table:bert-mlm} are included in Figure~\ref{fig:baseline-entropy}.

\subsection{Jensen-Shannon Divergence of Different Pre-trained Transformers} \label{sec:appendix:jsd}

\begin{figure*}[!h]
\centering
\subfigure[RealFormer]{\includegraphics[width=\textwidth, keepaspectratio]{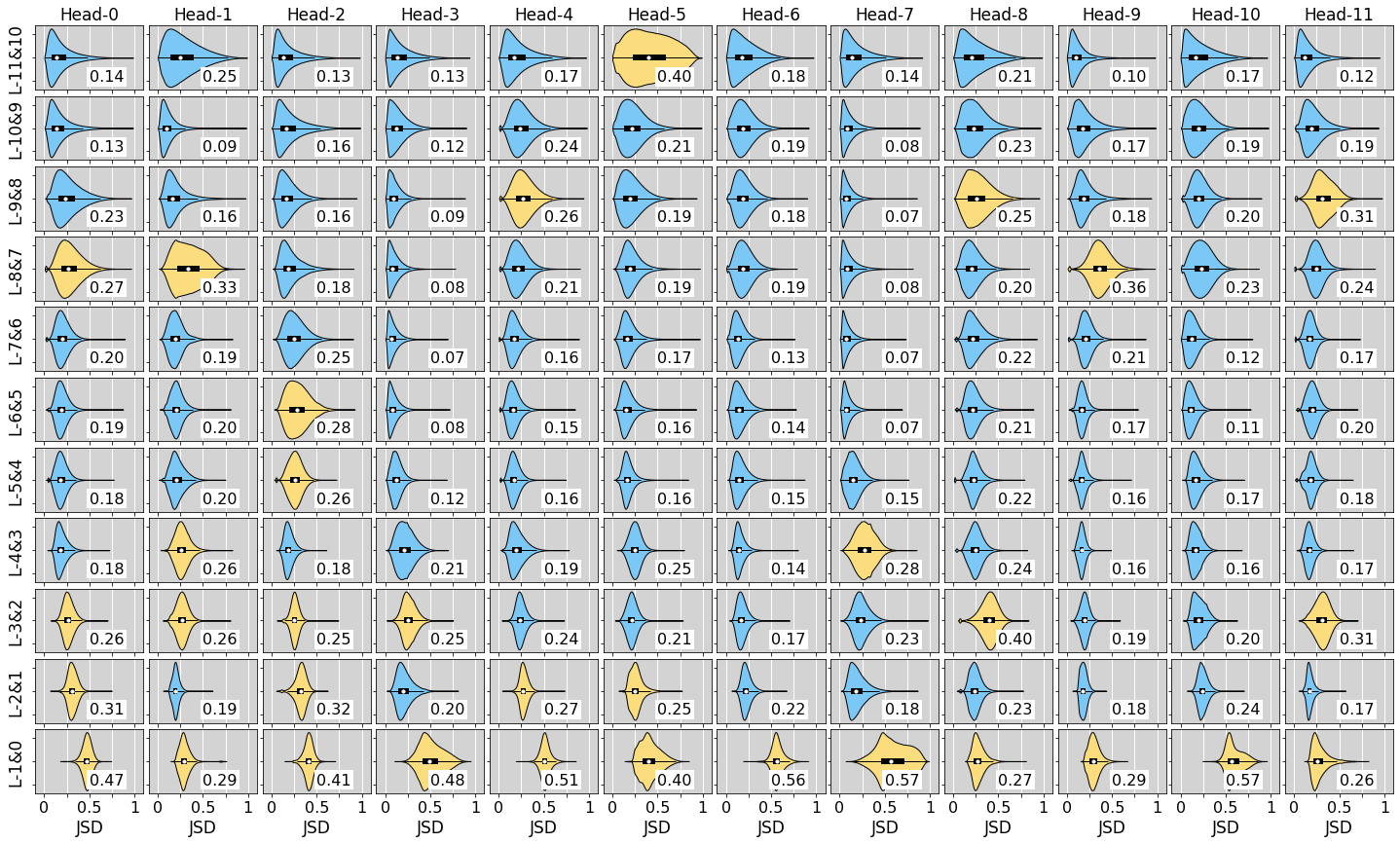}}
\subfigure[Post-LN]{\includegraphics[width=\textwidth,keepaspectratio]{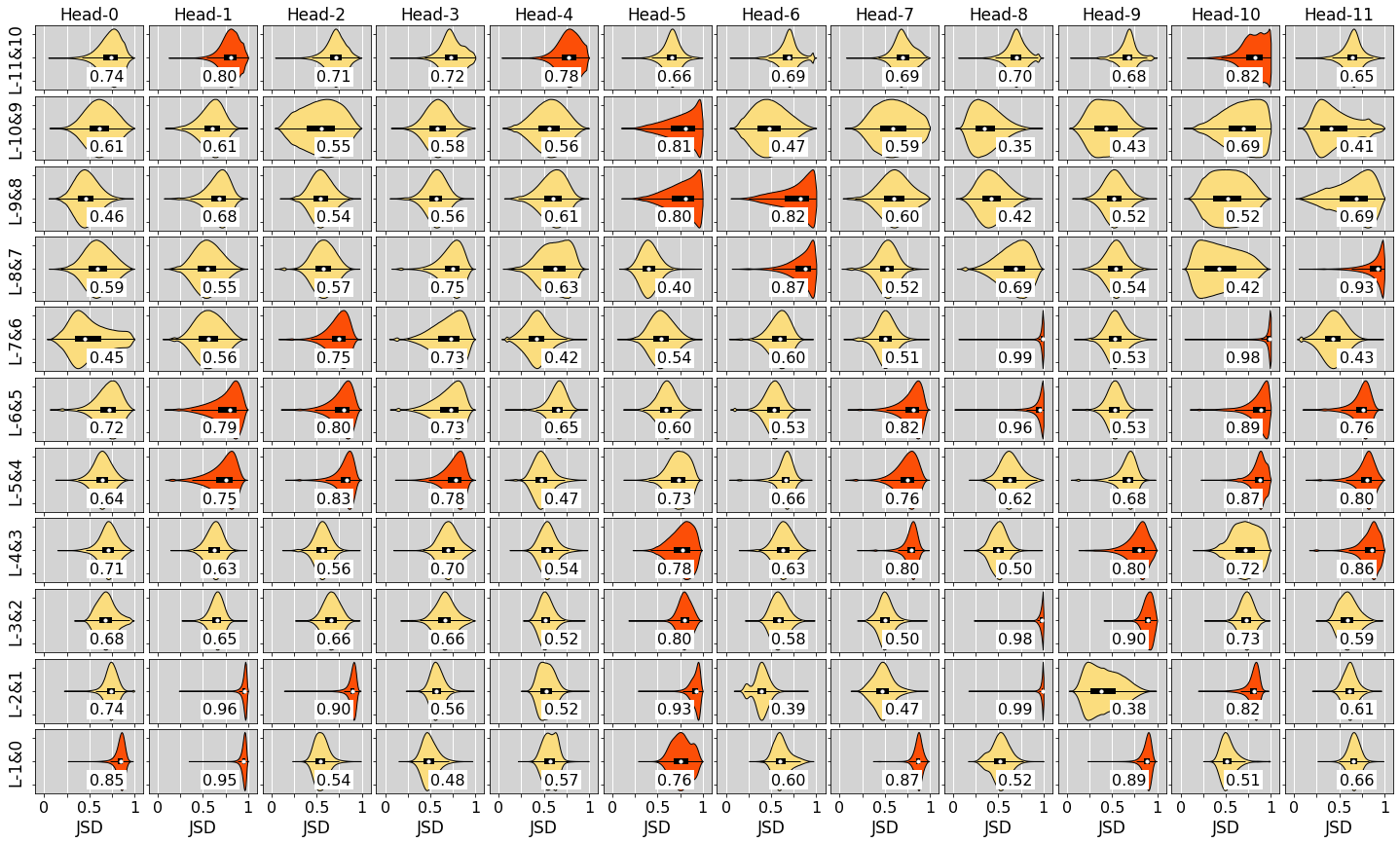}}
\caption{Distribution of Jensen-Shannon Divergence (JSD) of attention probabilities in (vertically) adjacent attention heads, \emph{i.e.}, $\text{JSD}(\text{head}^L_i, \text{head}^{L-1}_i)$. Based on 8,192 held-out examples using the pre-trained BERT-Base with \textbf{RealFormer} and \textbf{Post-LN} Transformer respectively (see Section~\ref{sec:pre-train}). Distributions are color-coded based on the median of JSDs: RED (median $>$ 0.75), YELLOW (0.25 $\le$ median $\le$ 0.75), BLUE (median $<$ 0.25). \emph{I.e.}, colder color means more ``similar'' attention heads across adjacent layers.}
\label{fig:jsd}
\end{figure*}

We use violin plots to show the Jensen-Shannon Divergence distributions of the pre-trained BERT-Base models with Post-LN and RealFormer from Table~\ref{table:bert-mlm} respectively (see Figure~\ref{fig:jsd}). 
Each row is a pair of adjacent layers in BERT-Base and each column is an attention head. 
Instead of computing one scalar value for each head pair, we show the full distribution based on the tokens in 8,192 held-out examples, \emph{i.e.}, each data point is the JSD between the attention probabilities of a token at these two heads.
For better legibility, we color code these plots to help distinguish head pairs with relatively ``similar'' attention (BLUE: median $<$ 0.25) and relatively ``distinct'' attention (RED: median $>$ 0.75) from the rest (YELLOW: 0.25 $\le$ median $\le$ 0.75).

Note that JSD results from Post-LN are used only as a reference; we expect them to be ``random'' because there is no correspondence between heads in adjacent layers for Post-/Pre-LN. Proof: An equivalent Post-/Pre-LN can be constructed by permuting the order of attention heads in a layer (and the corresponding variables).

\end{document}